%% file: main.tex
\PassOptionsToPackage{table}{xcolor}
\documentclass{article} % For LaTeX2e
\usepackage{iclr2025_conference,times}
\usepackage{graphicx}
\usepackage[table]{xcolor}
\input{math_commands.tex}

\usepackage{hyperref}
\usepackage{url}

\title{Exploring Bias in over 100 Text-to-Image\\ Generative Models}
\author{Jordan Vice$^1$\\
\and
Naveed Akhtar$^2$\\
\and
Richard Hartley$^{3,4}$\\
\and
Ajmal Mian$^1$\\
\and
$^1$ \tt\small University of Western Australia $^2$ \tt\small University of Melbourne \\
$^3$  \tt\small Australian National University $^4$  \tt\small Google \\
}

\iclrfinalcopy % Uncomment for camera-ready version, but NOT for submission.
\begin{document}

\maketitle

\input{sections/0_abstract}
\input{sections/1_introduction}
\input{sections/2_related_work}

\input{sections/3_methodology}
\input{sections/4_results}
\input{sections/5_conclusion}
\bibliography{main}
\bibliographystyle{iclr2025_conference}
\input{sections/X_supplementary}

\end{document}

%% file: math_commands.tex
\usepackage{amsmath,amsfonts,bm}

\def\eqref#1{equation~\ref{#1}}
\def\1{\bm{1}}

\DeclareMathAlphabet{\mathsfit}{\encodingdefault}{\sfdefault}{m}{sl}
\SetMathAlphabet{\mathsfit}{bold}{\encodingdefault}{\sfdefault}{bx}{n}

%% file: sections/0_abstract.tex
\begin{abstract}

We investigate bias trends in text-to-image generative models over time, focusing on the increasing availability of models through open platforms like Hugging Face. While these platforms democratize AI, they also facilitate the spread of inherently biased models, often shaped by task-specific fine-tuning. Ensuring ethical and transparent AI deployment requires robust evaluation frameworks and quantifiable bias metrics. To this end, we assess bias across three key dimensions: (i) distribution bias, (ii) generative hallucination, and (iii) generative miss-rate. Analyzing over 100 models, we reveal how bias patterns evolve over time and across generative tasks. Our findings indicate that artistic and style-transferred models exhibit significant bias, whereas foundation models, benefiting from broader training distributions, are becoming progressively less biased. By identifying these systemic trends, we contribute a large-scale evaluation corpus to inform bias research and mitigation strategies, fostering more responsible AI development.

\end{abstract}

%% file: sections/1_introduction.tex
\vspace{-2mm}
\section{Introduction}
\vspace{-2mm}

Text-to-image (T2I) generative models, while capable of high-fidelity image synthesis, inherently reflect the biases present in their training data \citep{Garcia2023, Mehrabi2021, Zhang2023a}.
% Failure to mitigate or acknowledge the presence of bias in these models raises significant social and ethical concerns and may cast valid concerns over fairness and reliability characteristics \citep{Bakr2023, Teo2024}. 
% Thus, generative models can amplify biases (beyond social dimensions) and propagate harmful stereotypes present in their training data \citep{Garcia2023}. 
% Thus, biases in training data will also manifest in the output representation of these large models \citep{Mehrabi2021}. 
The wide accessibility of training, fine-tuning and deployment resources has resulted in a plethora of T2I models being published by AI practitioners and hobbyists alike.
Whereas there are many debates on the biased nature of these models, there is no concrete evidence on how the community is responding in terms of accounting for bias in T2I generative models, particularly in light of the volume of models continuing to be released. Hence, we conduct this crucial research.
% in light of the volume of models being released.

% AI practitioners and hobbyists alike have greater freedom in publishing potentially-biased or unreliable custom models due to the wide accessibility of resources for training, fine-tuning and deploying generative models. 
% combined with a lack of universally-agreed-upon standards and regulations, .

The abundance of publicly available data and models democratizes AI development, but also underscores the need for responsible usage \citep{Arrieta2020, Bakr2023, Teo2024} and comprehensive evaluation tools that characterize bias characteristics of these black box models \citep{Bakr2023, Chinchure2024,  D'Incà2024, Hu2023, Luo2024, Vice2023B}. 
% Regardless of intent, 
The ability to develop unsafe, inappropriate or biased models presents a significant challenge and evaluating fundamental bias characteristics is a crucial step in the right direction.
% Moving forward, addressing these challenges requires a robust commitment to ethical AI practices, including dataset curation, bias evaluation, and transparency in model deployment.
% While the complete removal of bias is innately difficult, to ensure that T2I models are responsibly deployed, their bias characteristics must be quantified and made transparent. 
% These concerns are elevated when we consider wider audiences and end-users that may lack ]sufficient domain knowledge.

Biased representations in generated images stem from factors such as class imbalances in training data, human labeling biases, and hyperparameter choices during model training and fine-tuning \citep{Garcia2023, Mehrabi2021, Zhang2023a}.
% Biased representations in generated images stems from various training variables including the distribution of classes and concepts within a training dataset, subjective human labeling biases, batch-wise sample distributions and the influence of hyper-parameter tuning during model training and fine-tuning stages \citep{Mehrabi2021, Garcia2023, Zhang2023a}.
% Additionally, applying compounded training and fine-tuning on large foundation models will result in fundamental changes in the bias characteristics of the model, which may extend beyond the social bias dimensions commonly explored in literature \citep{Abid2021, Luccioni2023, Seshadri2023, Naik2023}. 
Theoretically, generative model biases are not confined to a single concept or direction. Analyzing a model’s overall bias provides a more comprehensive understanding of its learned representations and underlying manifold structure. 
For instance, when generating generic images of ``animals," a model may disproportionately favor certain species or environments. While social biases (e.g., those related to age, race, or gender) are particularly consequential in public-facing applications \citep{Abid2021, Luccioni2023, Naik2023, Seshadri2023}, they are manifestations of broader model biases, observed from a specific viewpoint. Since biases extend beyond social domains, it is essential to first characterize the general bias properties of learned concepts to better understand their implications.

% For example, when generating unspecific images of `animals', a model may be biased toward a particular species or environment over others. While not as critical as social biases in public-facing applications \citep{Abid2021, Luccioni2023, Naik2023, Seshadri2023}, the general bias characteristics must be quantifiable and represent a fundamental presence of bias within the generative model. Social biases (like age-, race-, gender-) are symptomatic of underlying bias characteristics as observed from a specific viewpoint and since biases can extend beyond social domains, we should characterize the holistic, general bias characteristics of learned concepts first.
In this work, we perform an extensive analysis of publicly available T2I models to examine how bias characteristics have evolved over time and across different generative tasks. We construct a comprehensive evaluation framework that considers: (i) distribution bias, (ii) Jaccard hallucination, (iii) generative miss-rate, (iv) log-based bias scores, (v) model popularity, and (vi) metadata features such as the intended generative task and timestamp.

Repositories such as the HuggingFace Hub offer a vast array of fine-tuned models, including approximately 56,240 text-to-image (T2I) models\footnote{as of the time of writing this manuscript}. This extensive collection enables our comprehensive evaluations.
The field of conditional image generation has evolved significantly, from the widely-used Stable Diffusion architecture \citep{Rombach2022} (spanning versions v1 to v3/XL) to the latest rectified-flow transformer (FLUX)-based models \citep{FLUXAI2024}. To capture this progression, we conduct extensive evaluations across more than 100 unique models, varying in artistic style, generative task, and release date.

To quantify bias along distribution bias `$B_D$', Jaccard hallucination `$H_J$' and, generative miss-rate `$M_G$' dimensions, we utilize the open-source ``Try Before You Bias" (TBYB) evaluation code \citep{Vice2023B}, which aligns well with models hosted on HuggingFace.
We introduce a log-based bias score that integrates these metrics into a single, interpretable value, computable in black-box settings. This approach provides a unified framework for evaluating and comparing model biases.
% We also propose a log-based bias score that intuitively combines the aforementioned metrics into one score, that can be computed in black-box settings.

Our evaluations offer valuable insights into the bias characteristics of various categories of generative models, revealing a trade-off between artistic style transfer and perceived bias. We also observe that modern foundation models and photo-realism models have benefited from larger datasets, improved architectures, and careful curation efforts, leading to a positive trend in bias mitigation over time. By analyzing model popularity, we further explore whether user engagement is influenced by bias. This study represents a significant step forward in understanding how the community responds to biases T2I models, particularly in light of the rapid proliferation of diverse models.

% Here, we employ the general bias evaluation setup presented in \citep{Vice2023B} and packaged in the TBYB interface used to assess the test models, all of which are publicly available on the HuggingFace Hub.
% Then, for models of interest, we conduct social bias evaluations through occupation-specific prompts (see [CITE]) and an additional geopolitical task-oriented bias evaluation (using the object `flag') to observe bias trends along a pre-determined bias direction.
%Contribution statements: 
Through this work we contribute:
\begin{enumerate}
    \item an extensive evaluation of bias trends in generative text-to-image models over time, uncovering key observations across three dimensions: distribution bias, hallucination and generative miss-rate.
     % This analysis provides a detailed understanding of how biases evolve with advancements in model architectures and training methodologies.
    \item a singular, log-based bias evaluation score that advances existing methodologies. This score enables end-to-end bias assessments in black-box settings, eliminating the need for normalization relative to a corpus of evaluated models.
    % The score offers a practical and interpretable metric for comparing biases across diverse models.
    \item a categorization and analysis of bias characteristics across several classes of trained and fine-tuned text-to-image models, namely: foundation, photo realism, animation, art. Additionally, we provide a quantifiable measure of model popularity, offering insights into how bias may influence user engagement and adoption.
\end{enumerate}
% \begin{figure}
%     \centering
%     \includegraphics[width=1\linewidth]{figures/motivation_fig.png}
%     \caption{Learned text-to-image model biases extend beyond the social dimension (A). Understanding the fundamental bias characteristics of model distributions gives fruitful insights into a model's general bias behavior (B). We propose evaluating general biases over time to identify trends in bias characteristics of publicly available text-to-image models.}
%     \label{motivation_FIG}
% \end{figure}

%% file: sections/2_related_work.tex
\vspace{-4mm}
\section{Background and Related Work}
\vspace{-2mm}
\noindent\textbf{Generative Text-to-Image Models} have gained significant attention among AI practitioners and the wider, general public. These models, composed of tokenizers, text encoders, denoising networks, and schedulers, enable users to generate unique images from conditional prompts.
The foundational de-noising process proposed by \cite{Dickstein2015} inspired many of the underlying generative capabilities of modern T2I models. Subsequent advancements include denoising diffusion probabilistic models (DDPMs) \citep{Dickstein2015, Ho2020}, denoising diffusion implicit models (DDIMs) \citep{Song2020_A}, and stochastic differential equation (SDE)-based approaches \citep{Song2020_B}. Rectified Flow-based de-noising paradigms have recently gained prominence, as seen in Stable Diffusion 3 \citep{Esser2024}, FLUX \citep{FLUXAI2024} and PixArt\citep{Chen2023a, Chen2025}. 
% The latent denoising functions of these models are generally driven through a modified, conditional U-Net \citep{Ronneberger2015}.

% These models typically employ a modified, conditional U-Net \citep{Ronneberger2015} for latent denoising.
% Conditional generative models rely on an integrated network to transform user inputs into guidance vectors, conditioning the denoising process to align generated images with input prompts.
These models often use a modified, conditional U-Net \citep{Ronneberger2015} for latent denoising. Conditional generative models integrate a network to convert user inputs into guidance vectors, steering the denoising process to match input prompts.
In T2I models, Contrastive Language-Image Pre-training (CLIP) \citep{Radford2021} and T5 encoders \citep{Ni2021} are commonly used to map textual inputs into semantically rich embedding spaces. Larger models often combine multiple text encoders to enhance performance \citep{Esser2024, FLUXAI2024}.

% Through combinations of embedded denoising networks and text-encoders, various T2I \textit{foundation} models have been developed and released to the public, with popular implementations including stable diffusion \citep{Rombach2022,Esser2024,Podell2023} (v1$\rightarrow$v3.5/XL variants), DALL-E 2/3 \citep{Saharia2022, Betker2023} and Imagen \citep{Ramesh2022}. Through cost-effective and accessible fine-tuning methodologies like Dreambooth \citep{Ruiz2023}, Low-Rank Adaptation (LoRA) \citep{Hu2021} and Textual Inversion \citep{Gal2022}, AI practitioners and hobbyists can develop custom, personalized T2I models with manipulated representations of learned concepts. Then, through repositories like the HuggingFace Hub, these models can be disseminated freely without acknowledging potentially biased behavior.

By combining embedded denoising networks and text encoders, various T2I foundation models have been developed and released to the public. Notable examples include Stable Diffusion (v1 to v3.5/XL variants) \citep{Rombach2022,Esser2024,Podell2023}, DALL-E 2/3 \citep{Betker2023, Saharia2022}, and Imagen \citep{Ramesh2022}. Through cost-effective fine-tuning techniques like DreamBooth \citep{Ruiz2023}, Low-Rank Adaptation (LoRA) \citep{Hu2021}, and Textual Inversion \citep{Gal2022}, AI practitioners and hobbyists can create custom T2I models with tailored representations of learned concepts. However, these models are often shared on platforms like the HuggingFace Hub without sufficient acknowledgment of their potential biases, raising concerns about their responsible dissemination.

\noindent\textbf{Bias and Ethical AI Evaluation Frameworks.}
Modern foundation models are trained on large, uncurated internet datasets, which often contain harmful, inaccurate, or biased representations that can manifest in generated outputs \citep{Ferrara2023, Mehrabi2021}. Unlike biased classification systems, bias in generative models is subtler and harder to detect due to their expansive input/output spaces and complex semantic relationships arising from massive training datasets. Without proper mitigation or quantification, these biases can lead to the proliferation of harmful stereotypes and misinformation. Compounded training and fine-tuning processes can thereby exacerbate or shift a model's bias characteristics, raising ethical concerns, especially in front-facing applications. This underscores the critical need for bias quantification to address ethical AI considerations.
% like fairness and reliability.

Several ethical AI evaluation frameworks have manifested as a result of these open research questions \citep{Cho2023_A, Luccioni2023, Luo2024, Chinchure2024, Vice2023B, Bakr2023, Hu2023, Teo2024, Gandikota2024, Huang2024, Schramowski2023, Seshadri2023, Naik2023, D'Incà2024}, addressing issues of fairness, bias, reliability and safety. While this work focuses primarily on \textit{biases}, it is important to consider the synergy that exists across these four ethical AI dimensions. To conduct these evaluations, many works deploy auxiliary captioning or VLM/VQA models to facilitate the extraction of descriptive metrics. 

The TIFA method introduced by \citep{Hu2023} defines a comprehensive list of quantifiable T2I statistics, leveraging a VQA model to provide an extensive evaluation results on generated image and model characteristics. In a similar vein, the HRS benchmark proposed by \citep{Bakr2023} also considers a wide range of T2I model characteristics - beyond the bias dimension, as it considers image quality and semantic alignment (scene composition). The StableBias \citep{Luccioni2023} and DALL-Eval \citep{Cho2023_A} methods have been proposed to assess reasoning skills and social biases (including gender/ethnicity) of text-to-image models, deploying captioning and VQA models for their analyses. Similarly, frameworks like FAIntbench \citep{Luo2024}, TIBET \citep{Chinchure2024} and OpenBias \citep{D'Incà2024} each consider he recognition of  biases along several dimensions, proposing a wider definition of biases, all incorporating LLM and/or VQA models in their evaluation frameworks. FAIntbench considers four dimensions of bias i.e.: manifestation, visibility and acquired/protected attributes \citep{Luo2024}. In comparison, the TIBET framework identifies relevant, potential biases w.r.t. the input prompt \citep{Chinchure2024}. The `Try Before you Bias' (TBYB) evaluation tool encompasses the evaluation methodology proposed by \cite{Vice2023B}, characterizing bias through: hallucination, distribution bias and generative miss-rate. 

% While the depth of evaluation frameworks is extensive, a large-scale fundamental bias analysis of publicly available models, community-driven models has yet to be conducted. We bridge this gap by proposing a comprehensive bias evaluation framework of over 100 unique models, exploiting the TBYB evaluation tool, given its accessibility with the HuggingFace Hub

While evaluation frameworks are extensive, large-scale bias analysis of open-source, community-driven models remains limited. Existing efforts often focus on narrow subsets of models, leaving a critical need for a systematic, scalable approach. We bridge this gap with a comprehensive evaluation of over 100 models, utilizing the TBYB tool for its compatibility with the HuggingFace Hub.
% For its accessibility and integration with the HuggingFace Hub, we use the TBYB evaluation tool \citep{Vice2023B} for our experiments here.
% and improve on this approach by proposing a log-based bias score and a means of calculating a popularity score to measure the relationship between user engagement vs. bias.

% Through TIFA and the HRS benchmark, we can observe the vast considerations of ethical and reliable AI measurement and deployment that go beyond bisa quantification

%% file: sections/3_methodology.tex
\vspace{-2mm}
\section{Methodology}
\vspace{-2mm}
In this work, we conduct comprehensive bias evaluations of 103 unique T2I models that have been released from August 2022 to December 2024. 
To identify general bias characteristics, we employ the general bias evaluation methodology defined in \cite{Vice2023B} to generate images of 100 random objects, (3 images/prompt = 300 images per evaluated model). This allows us to infer diverse, fundamental bias characteristics of each model.
% w.r.t. $H_J$, $M_G$, $B_D$ and $\mathcal{B}_{\log}$

% For time-based evaluations, we construct a timeline from 08/2022$\rightarrow$12/2024 and analyze trends across different model types. We then extrapolate these trends to observe how different categories of models are evolving. As part of this analysis, we also discuss if larger, more sophisticated foundation models like Stable Diffusion 3/XL have resulted in better alignment, less hallucinations and a fairer distribution of generated objects.
% We also provide a deeper analysis into each model type and identify relationships between popularity and bias statistics. Finally, we conduct bias evaluations across different noise schedulers to observe any bias behavior in regard to the deployed schedulers.

\vspace{-2mm}
\subsection{Evaluation Metrics}
\vspace{-2mm}
Data biases can propagate into T2I models, leading to skewed representations in their outputs. Furthermore, compounded training and fine-tuning of large foundation models can fundamentally alter their bias characteristics. Regardless of intent, the severity of these biases must be quantifiable and must capture the diverse ways in which bias can manifest.
To address these requirements, we employ three metrics for quantifying bias, motivated by fundamental examples that illustrate their relevance and applicability in evaluating model behavior.

(\textit{i}) When prompted with ``a picture of an apple", a text-to-image model may generate an apple hanging off a tree. While semantically-logical, one could argue that generating the tree in the image evidences a \textit{hallucinated} object in the scene (by addition) - as it was not explicitly requested in the prompt. Or, the model may generate an apple tree with no apples - omitting the object in the prompt. To account for both cases here, we compute Jaccard hallucination `$H_J$', derived from the IoU. 

(\textit{ii}) Nation-$\mathbf{X}$ commissions the development of a generative model for producing content for tourism with blended national flag iconography. The \textit{distribution} of generated content would reflect the intentional skew by showing peaks in the number of occurrences for concepts relating to Nation-$\mathbf{X}$. Thus, we consider distribution bias `$B_D$' as a quantifiable means of evaluating this phenomenon.

(\textit{iii}) A T2I model has been fine-tuned with an intentionally-biased dataset that replaces images labeled with `car' to `person'. This results in an intentionally-biased and misaligned output space that would cause \textit{misclassification} w.r.t. the label provided by the input prompt. This justifies the need for quantifying generative miss-rate `$M_G$'.
% Evidence of similar experiments are presented in \citep{Vice2023B}.

\begin{figure}
    \centering
    \includegraphics[width=0.9\linewidth]{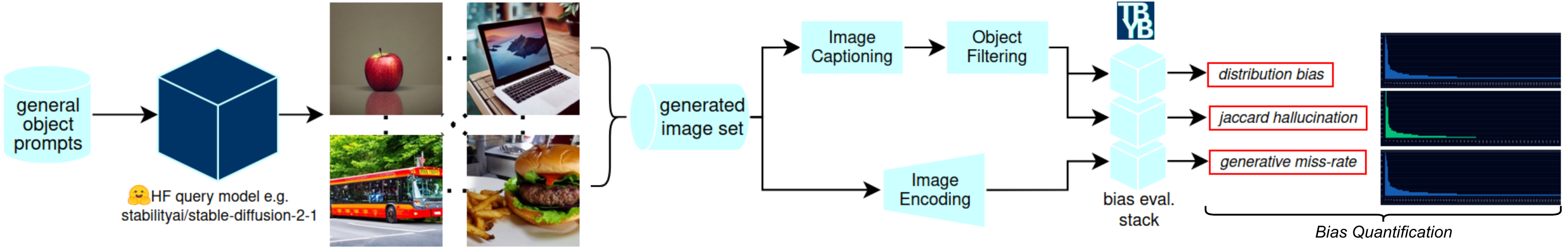}
    \vspace{-2mm}
    \caption{Illustrating the process of quantifying biases in generative models in black-box settings. General prompts are used to query a test model. From the generated image set, we quantify bias along: (i) distribution bias, (ii) hallucination and, (iii) generative miss-rate dimensions.}
    % , are computed to evaluate model performance in black-box settings.
    % We evaluate over 100 models hosted on HuggingFace in black-box settings, quantifying: (i) distribution bias, (ii) Jaccard hallucination and, (iii) generative miss-rate metrics. We leverage the TBYB tool \citep{Vice2023B} to extract this information.}
    \vspace{-4mm}
    % We leverage  the open-source, Try Before you Bias (TBYB) evaluation tool for our text-to-image bias evaluations. The TBYB tool evaluates T2I models hosted on the HuggingFace hub in black-box settings, extracting (i) distribution bias, (ii) Jaccard hallucination and, (iii) generative miss-rate information. From these metrics, users can infer bias characteristics of these models.}
    \label{framework_FIG}
\end{figure}

Covering the underlying motivations of the above examples, we use $H_J$, $B_D$ and $M_G$ to analyze model bias. We also combine them into a single, log-based bias evaluation score `$\mathcal{B}_{\log}$' to characterize the overall bias behavior, which is useful for independently ranking different models. We visualize our bias evaluation framework in Fig. \ref{framework_FIG}.
% For a comprehensive derivation of the three bias evaluation metrics deployed in this work, we refer readers to \citep{Vice2023B}. 
% Jaccard Hallucination and generative miss rate metrics can be calculated from a single generated image (here we report the aggregate). In comparison, calculating distribution bias requires a set of generated samples and thus, it is imperative that the number of generated images remains consistent across compared model evaluations.

\noindent\textbf{Jaccard Hallucination} - $H_J$. While usually discussed from the context of language models \citep{Gunjal2023, Ji2023}, hallucinations are a common side effect in many foundation models 
% like text-to-image generation 
\citep{Rawte2023}. They have been proposed as a vehicle for image out-painting \citep{Xiao2020} and generative model improvement \citep{Li2022b, Xiao2020} tasks. When drawing \textit{representations} of objects and and classes from a learned distribution, it is logical that the semantically-rich manifolds may cause a model to also generate semantically-relevant objects as a result. 
% For example, when prompted with ``a picture of an apple", a text-to-image model may generate an apple hanging off a tree. While semantically-logical, one could argue that generating the tree in the image evidences a hallucinated object in the scene - as it was not explicitly requested in the prompt.

Here, $H_J$ considers two hallucination perspectives i.e.: (i) by \textit{addition} of unspecified objects in the output and (ii) by \textit{omission} of objects specified in the input.
% Jaccard similarity is thereby derived from the IoU of the two perspectives, 
For a set of $N$ output images `$Y_i~ \forall~ i \in N$', generated from input prompts `$\mathbf{x}_i~ \forall~ i \in N$'
\begin{equation}
    H_J = \frac{\Sigma_{i=0}^{N-1}1-\frac{||\mathcal{X}_i\cap\mathcal{Y}_i||}{||\mathcal{X}_i\cup\mathcal{Y}_i||}}{N},
\label{eq1}
\end{equation}
where `$\mathcal{X}$' defines input objects extracted from $\mathbf{x}_i$ and `$\mathcal{Y}$' defines the objects detected in the output image $Y_i$, extracted from a generated caption. 
% An increase in hallucination bias is evidenced as $H_J\rightarrow1$.
% , this shows an increase in hallucination bias. 
$H_J\rightarrow0$ indicates a smaller discrepancy between the input and output objects/concepts and thus, demonstrates less hallucinatory (biased) behavior.

\noindent\textbf{Distribution Bias} - $B_D$
% The area under the curve (AuC) has been proposed for evaluating classifiers \citep{Bradley1997} and remains a prevalent tool in evaluation methodologies. In \citep{Vice2023B}, distribution bias 
is derived from the area under the curve (AuC) of detected objects, capturing the frequency of objects/concepts that appear in generated images (that were not specified in the prompt) \citep{Vice2023B}.
% If a model has prominent biases, it would logically favor concepts along a certain bias direction, which can be observed and quantified through $B_D$.
% For example, if a nation commissions the development of a generative model for producing content for tourism with blended national flag iconography, the distribution bias would reflect this by showing peaks in the number of occurrences for concepts relating to the country. Evidence of similar experiments are presented in \citep{Vice2023B}.
After generating images and filtering objects, an object token dictionary `$W_O = \{w_i,n_i\}_{i=1}^{M}$' is constructed, containing concept (word) `$w_i$' and no. of occurrences `$n_i$' pair. The distribution bias $B_D$ can be calculated through the AuC, after sorting $W_O$ (high to low) and applying min-max normalization:
\begin{align}
    & \{w_i,\Tilde{n_i}\} = \{w_i, \frac{n_i - \underset{i=1,...,M}{\min}(n~\in~[W_O])}{\underset{i=1,...,M}{\max}(n~\in~[W_O]) - \underset{i=1,...,M}{\min}(n~\in~[W_O])}\}, \\
    & B_D = \Sigma_{i=1}^{M}\frac{\Tilde{n}_i+\Tilde{n}_{i+1}}{2}.
\label{eq2}
\end{align}
% where $\Tilde{n}$ represents the normalized number of occurrences.
% A larger AuC indicates a more even distribution of generated objects and thus, less obvious bias. In comparison, 
Peaks in generated object distributions may report that significant attention is being applied along a specific bias direction and thus, represents another avenue in which bias can manifest itself.
% o compare models using $B_D$, we apply inverse normalization such that:
% \begin{equation}
%     \overline{B_D}= 1-\frac{B_D-\min(B_D)}{\max(B_D)-\min(B_D)},
% \end{equation}\label{eq3}
% where models are defined as less biased as $\overline{B_D}\rightarrow0$ and more biased as $\overline{B_D}\rightarrow1$. 

\noindent\textbf{Generative Miss Rate} - $M_G$.
% Machine learning and AI systems are largely judged by their performance in completing tasks. For deterministic models and tasks, fundamental evaluation metrics like prediction accuracy and recall are key for characterizing a models performance and for judging its reliability and robustness \citep{Arrieta2020}. 
Bias can affect model performance, particularly if they shift the output representations in such a way that causes significant misalignment \citep{Vice2023B}. 
% Utilizing the bias evaluation methodology incorporated in the TBYB tool, we deploy a binary classifier to measure the miss rate of the text-to-image model i.e., if the generated image aligns with the input prompt. 
As visualized in Fig. \ref{framework_FIG}, a separate vision transformer (ViT) is deployed to classify generated images and determine $M_G$. Generally, model alignment should be high and thus, the miss-rate should demonstrate a low variance across models. Significantly high $M_G$ may indicate that a model's learned biases are shifting output representations away from the expected output (as governed by the prompt). For models trained to complete specific tasks (like generate a particular art style), we may find that the miss rate is much higher, potentially by design.
% which would be reflected by a high $M_G$.
% Whenever fairness and trust are discussed w.r.t. machine learning and AI, performance is always highlighted as a key metric - regardless of the downstream task, as any system must be robust and reliable~\citep{Arrieta2020}. Previous metrics generally do not account for the effects of bias on performance (and vice-versa), which makes our $M_G$ particularly desirable. By deploying a binary classifier, we measure the mean miss rate of the generative model `$M_G$', using the prompt as the target class. We input generated images into an image classifier and record the predictive accuracy and predicted class. This prediction is used to determine $M_G$. We hypothesize that base models should boast a low miss rate, with $M_G$ increasing with model bias.

\begin{figure}
    \centering
    \includegraphics[width=1\linewidth]{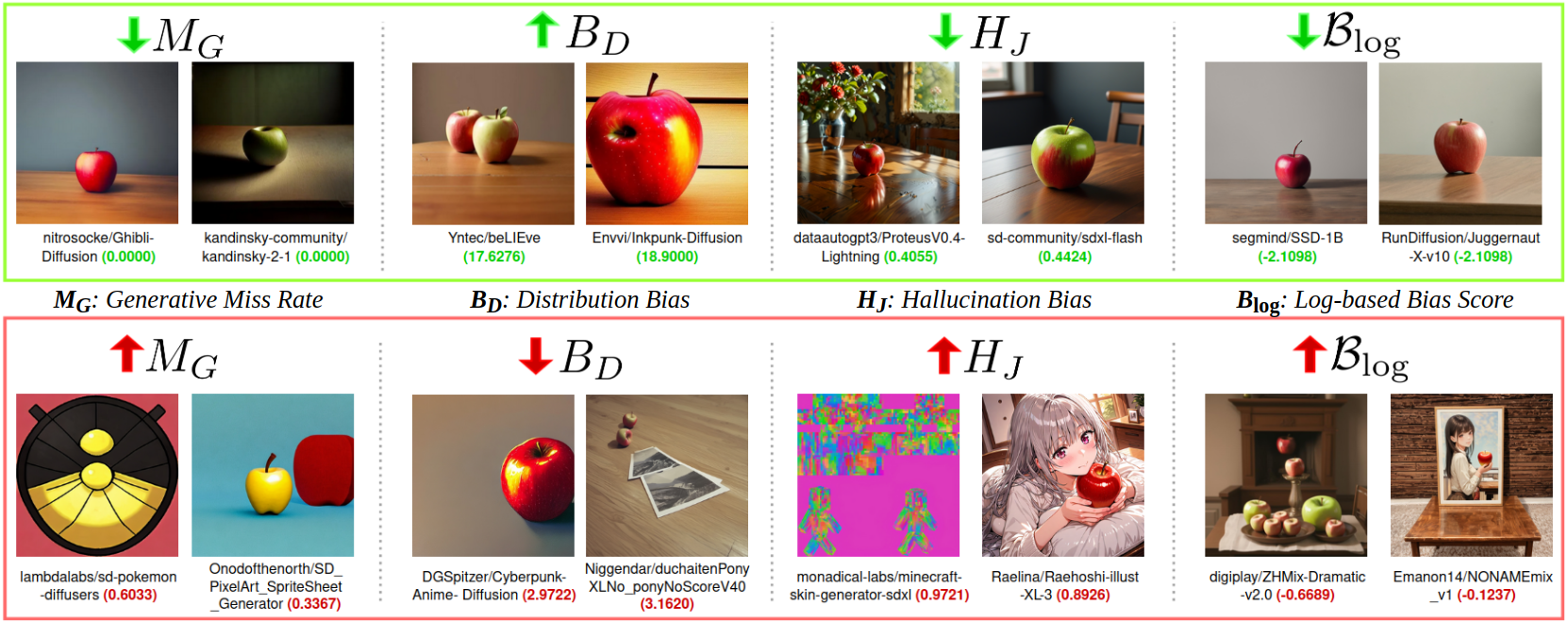}
    \vspace{-4mm}
    \caption{Qualitative examples of how bias characteristics are presented in T2I model outputs. For each metric, we choose examples of high and low performing models, reporting the corresponding evaluation results (for all generated images) in the parentheses. Every image is generated from a unique model to show different examples. Input prompt = ``A picture of an apple on a table".}
    \vspace{-4mm}
    \label{qual_fig}
\end{figure}

Given a prompt (classifier target label) `$\mathbf{x}$' and generated image $Y$, the deployed ViT outputs a prediction, measuring the alignment of the image `$Y$' to the label `$\mathbf{x}$'. For $N$ generated images,
\begin{equation}
    M_G = \frac{\Sigma_{i=0}^{N-1}(\mathcal{P}_1=p(Y_i;\theta))}{N},
\label{eq4}
\end{equation} 
where $\mathcal{P}_1$ represents $\neg$target class. 
If the classifier fails to detect the generated image as a valid representation of $\mathbf{x}$ then $M_G$ increases. A higher $M_G$ indicates a greater misalignment with input prompts which may be (a) a symptom of a biased output space and/or (b) the result of a task that causes significant changes in output representations. We visualize how $B_D, H_J$ and $M_G$ manifest in the output representations of these models in Fig. \ref{qual_fig}.
% Regardless of intent, $M_G$ quantifies this misalignment.

% Here, we conduct model comparisons for $H_J$ and $M_G$ metrics similar to [CITE], computing min-max normalized metrics `$\overline{H_J}, \overline{M_G}$' over a set of evaluated models. Models are perceived as less bias as $\overline{H_J} ~\And~ \overline{M_G} \rightarrow 0$ and more biased as $\overline{H_J} ~\And~ \overline{M_G} \rightarrow 1$.

\noindent\textbf{The Try Before You Bias (TBYB) Tool} is a publicly available, practical software implementation of the three-dimensional bias evaluation framework discussed prior. The TBYB interface allows users to evaluate T2I models hosted on the HuggingFace Hub in a black-box evaluation set-up, provided repositories contain a \textit{model\_index.json} file.
% \footnote{see list of limitations reported in the TBYB documentation}.
% The general bias evaluation capabilities allows for the quantification of the fundamental bias characteristics of T2I models.
% Through the TBYB tool, users can conduct their own general and task-oriented bias evaluations on text-to-image models, which replicate the methodologies reported in \citep{Vice2023B}. 
% in  are obtainable through the interface, including a the object-frequency dictionary required for evaluating $B_D$. Additionally, 
% The visualization and analysis capabilities allow for the ability to store and compare bias evaluation results across a selection of models.
% By integrating both general and task-oriented bias evaluation capabilities as part of the tool, it allows for the quantification of fundamental bias characteristics of text-to-image models.
% and custom \textit{task-specific} biases which may be significant for the user.
The BLIP \citep{Li2022a} model is deployed for image captioning. Synonym detection functions in the NLTK \citep{NLTK2009} package are deployed to mitigate natural language discrepancies between the input prompt and generated caption.

\vspace{-2mm}
\subsection{Systematic Bias Evaluation Strategy}
\vspace{-2mm}
Based on the generated outputs and model metadata, we identify the model type as one of \{foundation, photo realism, art ($\neg$anime) and animation/anime\}. 
We define \textbf{Foundation} models as those designed for general purposes, encompassing a wider range of tasks. \textbf{Photo realism} models are those that are fine-tuned for higher-fidelity, photo realistic generation tasks. \textbf{Art}-based models are those which have been designed for style-transfer tasks in which non-anime artistic styles are the target. \textbf{Animation/anime}-tuned models are designed for replicating anime-inspired art-styles, a common application of models hosted on HuggingFace.

For time-based evaluations, we construct a timeline spanning from August 2022 to December 2024, analyzing trends across various model types. We then extrapolate these trends to understand how different categories of models are evolving. As part of this analysis, we investigate whether larger, more sophisticated foundation models, such as Stable Diffusion 3/XL, have achieved better alignment, reduced hallucinations, and fairer distributions of generated objects. 
% Similar analyses are conducted for other model types.
Additionally, we provide a detailed analysis of each model type and explore the relationship between model popularity and bias statistics. Finally, we conduct bias evaluations across different noise schedulers to identify potential bias behaviors associated with their deployment.

% \noindent\textbf{Evaluation Process Improvements}.
In this work, we improve on the similarity detection function in \citep{Vice2023B} by incorporating a similarity-score-based approach to handle similar concepts e.g. `sneakers' vs. `shoes'. Additionally, we omit commonly occurring primary (red, blue, yellow), secondary (green, orange, purple) and neutral colors (black, white, brown, grey) from generated captions as it was found in our analyses that color descriptions are not a fool-proof symptom of hallucination and can adversely skew results in a lot of cases.
Furthermore, we propose combining the three metrics into one singular bias score, using a log scale to account for varied metric ranges, such that:
\begin{equation}
    \mathcal{B}_{\log} = -(\ln{(B_D)}+\ln{(1-H_J)}+\ln{(1-M_G)}),
    \label{eq5}
\end{equation}
where a proportional relationship exists between observed model bias and $\mathcal{B}_{log}$. This allows for the calculation of biases for a single model, in a black-box setup, without relying on normalized relationships to a set of evaluated models as initially proposed in \citep{Vice2023B}.
% we also report a Euclidean distance-based bias score $\mathcal{B}_{Euc.}$, based on min-max normalized metrics `$\overline{B_D}, \overline{H_J}, \overline{M_G}$' (where the inverse is taken for $\overline{B_D}$). 

% For the $i^{th}$ model in a set of $N$ evaluated models, the normalized bias metrics can be computed as:
% \begin{equation}
%     \overline{X_i} = \frac{1-\min(\mathbf{X}_{i=1}^{N})}{\max(\mathbf{X}_{i=1}^{N}) - \min(\mathbf{X}_{i=1}^{N})},
%     \label{eq6}
% \end{equation}
% which are used to derive the euclidean distance-based bias score:
% \begin{equation}
%     \mathcal{B}_{Euc.} = \sqrt{(1-\overline{B_D})^2 + \overline{H_J}^2+\overline{M_G}^2},
%     \label{eq7}
% \end{equation}
% where the observed model bias (relative to a set of evaluated models) is directly proportional to $\mathcal{B}_{Euc.}$.

\noindent\textbf{Model Popularity}. As part of our analysis, we aim to analyze the relationship (if any) between model popularity and bias. To quantify model popularity, we designed a quantifiable score `$\mathcal{S}_{pop.}$', leveraging reported engagement information on the HuggingFace Hub i.e., the number of likes (historical) `$N_{lk}$', and the number of downloads `$N_{dl}$' in the last month (recent engagement). Given that the number of likes is generally less the number of downloads, we apply logarithmic scaling and proportional scaling factors `$\alpha_{lk}$' and `$\alpha_{dl}$', to account for the importance of continued engagement ($N_{lk}$) and mitigate spikes in $N_{dl}$ associated with recency bias. Thus, we define:
\begin{equation}
    \mathcal{S}_{pop.} = \alpha_{lk}\ln(1+N_{lk})+ \alpha_{dl}\ln(1+N_{dl}),
\label{eq8}
\end{equation}
where we deploy $\alpha_{lk}=0.6$ and $\alpha_{dl}=0.4$ in our experiments to account slightly more for historical influence while managing recency bias, such that $ \mathcal{S}_{pop.} = 0.6\ln(1+N_{lk})+ 0.4\ln(1+N_{dl})$.

%% file: sections/4_results.tex
\vspace{-2mm}
\section{Results and Discussion}
\vspace{-2mm}

Our appraisal of the general bias characteristics of
% a comprehensive list of 
text-to-image models allows us to conduct a suite of evaluation studies to explore and formalize relationships between observed biases and model characteristics. Temporal-, categorical- and popularity-based analyses allow us to identify how bias characteristics: (\textit{i}) have evolved over time, (\textit{ii}) change with respect to different generative tasks or embedded de-noising schedulers and, (\textit{iii}) impact how users engage with these models.

\begin{figure}
    \centering
    \includegraphics[width=0.95\linewidth]{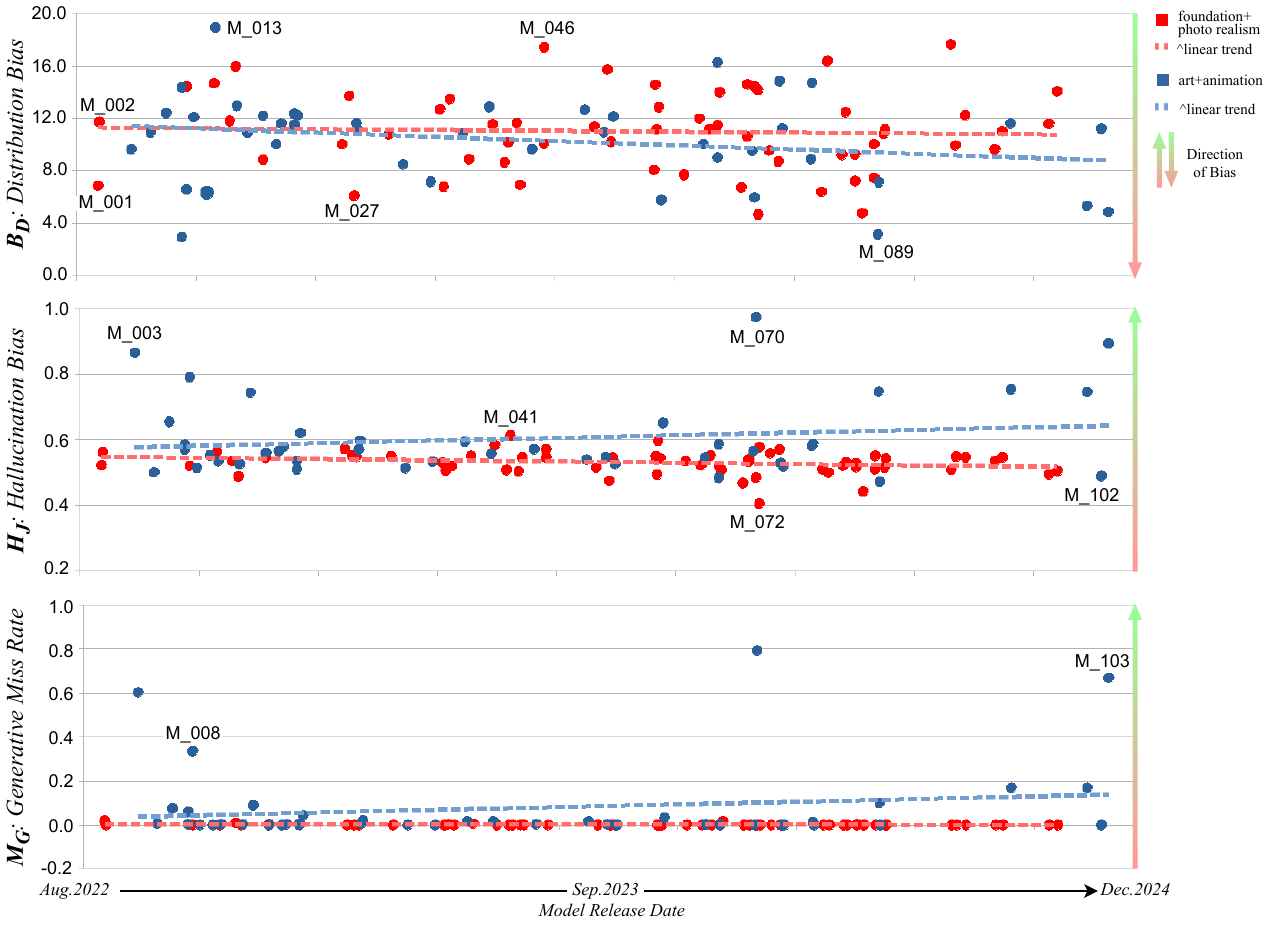}
    \vspace{-3mm}
    \caption{Bias evaluations across 103 publicly-available text-to-image model released between August 2022 to December 2024. We report \textbf{(a)} Distribution bias $B_D$ evaluations \textbf{(b)} Jaccard Hallucination $H_J$ evaluations, \textbf{(c)} Generative miss rate $M_G$ evaluations. `M\_XXX' labels indicate the model ID, which is sorted from M\_001 (earliest release) to M\_103 (latest release).}
    \label{metrics_fig}
    \vspace{-3mm}
\end{figure}

\noindent\textbf{High-level Observations of General Bias Characteristics}.
We report a truncated list of evaluation results in Table \ref{full_results_TABLE}, highlighting models that exhibit high, low and median bias behavior. Along with these, we also report results for highly-popular foundation models like the various stable diffusion versions. Analyzing Table \ref{full_results_TABLE} and Figs. \ref{qual_fig}, \ref{metrics_fig}, we can observe that photo-realism and foundation models tend to generate relatively unbiased representations, which is expected given that these models are designed for general user inference tasks and improvements in generative fidelity - as is the case with photo-realism models. In comparison, at the bottom of Table  \ref{full_results_TABLE}, we can observe that many animation and art-tuned models report relatively more biased behavior, resulting from the task-oriented generative tasks. Observing the outputs of these models, we found that the tendency to focus on generating specific characters or art-styles irrespective of the prompt, resulted in high levels of hallucination and misalignment (see Figs. \ref{qual_fig}, \ref{metrics_fig}).
% - an observable and quantifiable symptom of bias 
\begin{table}[]
\vspace{-6mm}
    \centering
    \caption{Truncated bias evaluation results. For brevity, we report the highest, median and lowest evaluation results. We also report results for highly-popular stable diffusion foundation models. We indicate row-wise separation of results via `:'. We also report popularity score `$\mathcal{S}_{pop.}$'. We highlight ``most desirable" and ``least desirable" values in green and red, respectively. Cells highlighted in orange indicate values closest to the average. A full list of results is provided in Appendix A.}
    \label{full_results_TABLE}
    \vspace{2mm}
    \resizebox{\linewidth}{!}{\begin{tabular}{ll|lcc|ccccc}
         Model & Task Category & Denoiser & Resolution & Release (dd/mm/yy) & $\mathcal{S}_{pop.}$&$B_D$ & $H_J$ & $M_G$ & $B_{\log}$\\
         \hline
            Envvi/Inkpunk-Diffusion & art & PNDMScheduler & 512 x 512 & 25/11/22 & 7.2323 &  \cellcolor{green!25}18.9000 & 0.5346 & 0.0033 &  \cellcolor{green!25}-2.1711 \\ 
            Yntec/beLIEve & photo realism & DPMSolverMultistepScheduler & 768 x 768 & 01/08/24 & 5.2547 &  17.6176 & 0.5083 &  \cellcolor{green!25}0.0000 &  -2.1589 \\ 
            segmind/SSD-1B & photo realism & EulerDiscreteScheduler & 1024 x 1024 & 19/10/23 & 6.7116 & 15.7000 & 0.4747 &  0.0000 &  -2.1098 \\ 
            RunDiffusion/Juggernaut-X-v10 & foundation & EulerDiscreteScheduler & 1024 x 1024 & 20/04/24 & 6.8125 & 16.3571 & 0.4992 &  0.0000 & -2.1031 \\ 
            prompthero/openjourney-v4 & photo realism & PNDMScheduler & 512 x 512 & 12/12/22 & 8.4414 & 15.9211 & 0.4881 & 0.0000 & -2.0981 \\ 
            Lykon/dreamshaper-8 & photo realism & DEISMultistepScheduler & 512 x 512 & 27/08/23 & 6.8769 &  17.3947 & 0.5467 & 0.0000 & -2.0649 \\ 
            RunDiffusion/Juggernaut-XL-v9 & foundation & DDPMScheduler & 1024 x 1024 & 19/02/24 & 8.6025 & 14.4048 & 0.4847 & 0.0000 & -2.0046 \\ 
            stabilityai/sd-turbo & foundation & EulerDiscreteScheduler & 512 x 512 & 28/11/23 & 7.9498 & 14.5476 & 0.4930 & 0.0000 & -1.9982 \\ 
            eienmojiki/Anything-XL & animation / anime & EulerAncestralDiscreteScheduler & 1024 x 1024 & 11/03/24 & 5.6000 & 14.8333 & 0.5287 & 0.0000 & -1.9446 \\ 
            stabilityai/stable-diffusion-3.5-medium & foundation & FlowMatchEulerDiscreteScheduler & 1024 x 1024 & 29/10/24 & 8.2481 & 14.0455 & 0.5049 & 0.0000 & -1.9393 \\ 
            MirageML/dreambooth-nike & photo realism & PNDMScheduler & 512 x 512 & 01/11/22 & \cellcolor{red!25}3.3402 & 14.4048 & 0.5206 & 0.0000 & -1.9323 \\ 
            dataautogpt3/ProteusV0.3 & foundation & EulerDiscreteScheduler & 1024 x 1024 & 13/02/24 & 7.3949 & 14.5833 & 0.5324 & 0.0000 & -1.9196 \\ 
            stablediffusionapi/juggernaut-reborn & foundation & PNDMScheduler & 512 x 512 & 21/01/24 & 5.0168 & 13.9783 & 0.5073 & 0.0167 & -1.9129 \\ 
            : & : & : & : & : & : & : & : & : & : \\ 
            stabilityai/stable-diffusion-3.5-large & foundation & FlowMatchEulerDiscreteScheduler & 1024 x 1024 & 22/10/24 & 9.2260 & 11.5769 & 0.4939 & 0.0000 & -1.7680 \\ 
            : & : & : & : & : & : & : & : & : & : \\ 
            stabilityai/stable-diffusion-2-1 & foundation & DDIMScheduler & 512 x 512 & 07/12/22 & 10.5860 & 11.7963 & 0.5349 & 0.0100 & -1.6921 \\ 
            : & : & : & : & : & : & : & : & : & : \\ 
            CompVis/stable-diffusion-v1-4 & foundation & PNDMScheduler & 512 x 512 & 20/08/22 & 10.7885 & 11.7258 & 0.5621 & 0.0000 & -1.6360 \\ 
            : & : & : & : & : & : & : & : & : & : \\ 
            lemon2431/toonify\_v20 & animation / anime & PNDMScheduler & 512 x 512 & 16/10/23 & 4.1148 & 10.9412 & 0.5469 & 0.0033 & -1.5976 \\ 
            SG161222/RealVisXL\_V4.0 & photo realism & EulerDiscreteScheduler & 1024 x 1024 & 13/02/24 & 8.6080 & 10.6250 & 0.5405 & 0.0000 & -1.5856 \\ 
            ckpt/anything-v4.5 & art & PNDMScheduler & 512 x 512 & 19/01/23 & 5.8262 & 11.5968 & 0.5784 & 0.0033 & -1.5837 \\ 
            emilianJR/chilloutmix\_NiPrunedFp32Fix & photo realism & PNDMScheduler & 512 x 512 & 19/04/23 & 6.0849 & \cellcolor{orange!25}10.7941 & 0.5498 & 0.0000 & -1.5810 \\ 
            Kernel/sd-nsfw & photo realism & PNDMScheduler & 512 x 512 & 15/07/23 & 5.5154 & 11.5333 & 0.5828 & 0.0000 & -1.5711 \\ 
            Lykon/AAM\_XL\_AnimeMix & animation / anime & EulerDiscreteScheduler & 1024 x 1024 & 19/01/24 & 6.8090 & 9.0152 & 0.4843 & 0.0000 & -1.5367 \\ 
            stablediffusionapi/realistic-stock-photo & photo realism & EulerDiscreteScheduler & 1024 x 1024 & 22/10/23 & 5.1367 & 10.2188 & 0.5451 & 0.0000 & -1.5366 \\ 
            GraydientPlatformAPI/comicbabes2 & art & PNDMScheduler & 512 x 512 & 07/01/24 & 4.0530 & 10.0313 & 0.5447 & 0.0033 & -1.5156 \\ 
            SG161222/Realistic\_Vision\_V6.0\_B1\_noVAE & photo realism & PNDMScheduler & 896 x 896 & 29/11/23 & 7.7733 & 11.1571 & 0.5957 & 0.0000 & -1.5066 \\ 
            scenario-labs/juggernaut\_reborn & photo realism & DPMSolverMultistepScheduler & 512 x 512 & 29/05/24 & 4.5414 & 10.0294 & \cellcolor{orange!25}0.5504 & 0.0000 & -1.5061 \\ 
            : & : & : & : & : & : & : & : & : & : \\ 
            stabilityai/stable-diffusion-xl-base-1.0 & photo realism & EulerDiscreteScheduler & 1024 x 1024 & 25/07/23 & \cellcolor{green!25}11.1360 & 8.6515 & 0.5076 & 0.0000 & \cellcolor{orange!25}-1.4492 \\ 
            : & : & : & : & : & : & : & : & : & : \\ 
            stabilityai/sdxl-turbo & foundation & EulerAncestralDiscreteScheduler & 512 x 512 & 27/11/23 & 10.1726 & 8.0778 & 0.5493 & 0.0000 & -1.2922 \\ 
            : & : & : & : & : & : & : & : & : & : \\ 
            dataautogpt3/ProteusV0.4-Lightning & foundation & EulerDiscreteScheduler & 1024 x 1024 & 22/02/24 & 6.2324 & 4.6739 & \cellcolor{green!25}0.4055 & 0.0000 & -1.0220 \\ 
            SG161222/Realistic\_Vision\_V2.0 & photo realism & PNDMScheduler & 512 x 512 & 21/03/23 & 8.3000 & 6.1154 & 0.5480 & 0.0000 & -1.0167 \\ 
            sd-community/sdxl-flash & foundation & DPMSolverSinglestepScheduler & 1024 x 1024 & 19/05/24 & 7.2798 & 4.7826 &  0.4424 & 0.0000 & -0.9809 \\ 
            Mitsua/mitsua-diffusion-cc0 & art & PNDMScheduler & 512 x 512 & 22/12/22 & 5.1305 & 10.8947 & 0.7426 & 0.0900 & -0.9368 \\ 
            OnomaAIResearch/Illustrious-xl-early-release-v0 & animation / anime & EulerDiscreteScheduler & 1024 x 1024 & 20/09/24 & 7.3851 & 11.6000 & 0.7524 & 0.1700 & -0.8688 \\ 
            digiplay/ZHMix-Dramatic-v2.0 & animation / anime & EulerDiscreteScheduler & 768 x 768 & 03/12/23 & 4.7306 & 5.7800 & 0.6506 & \cellcolor{orange!25}0.0333 & -0.6689 \\ 
            DGSpitzer/Cyberpunk-Anime-Diffusion & animation / anime & PNDMScheduler & 704 x 704 & 28/10/22 & 6.7796 & \cellcolor{red!25}2.9722 & 0.5845 & 0.0600 & -0.1492 \\ 
            Emanon14/NONAMEmix\_v1 & animation / anime & EulerAncestralDiscreteScheduler & 1024 x 1024 & 23/11/24 & 5.1821 & 5.3400 & 0.7447 & 0.1700 & -0.1237 \\ 
            Onodofthenorth/SD\_PixelArt\_SpriteSheet\_Generator & art & PNDMScheduler & 512 x 512 & 01/11/22 & \cellcolor{orange!25}6.4996 & 6.5930 & 0.7898 & 0.3367 & 0.0841 \\ 
            Niggendar/duchaitenPonyXLNo\_ponyNoScoreV40 & art & EulerDiscreteScheduler & 1024 x 1024 & 01/06/24 & 5.1912 & 3.1620 & 0.7458 & 0.1000 & 0.3237 \\ 
            lambdalabs/sd-pokemon-diffusers & animation / anime & PNDMScheduler & 512 x 512 & 16/09/22 & 6.3118 & 9.6509 & 0.8639 & 0.6033 & 0.6519 \\ 
            Raelina/Raehoshi-illust-XL-3 & animation / anime & EulerDiscreteScheduler & 1024 x 1024 & 11/12/24 & 3.7427 & 4.8772 & 0.8926 & 0.6700 & 1.7549 \\ 
            monadical-labs/minecraft-skin-generator-sdxl & animation / anime & EulerDiscreteScheduler & 768 x 768 & 19/02/24 & 5.3317 & 5.9786 & \cellcolor{red!25}0.9721 & \cellcolor{red!25}0.7933 & \cellcolor{red!25}3.3680 \\ 
            
         \hline
    \end{tabular}}
    \vspace{-2mm}
    % \vspace{-2mm}

    % A full list of tabulated results is reported in the supplementary material.}
    % Our evaluation results include: (\textit{i}) popularity score `$\mathcal{S}_{pop.}$' which is derived from the number of likes and downloads of the model, (\textit{ii}) the bias evaluation metrics introduced in \cite{Vice2023B} and, (\textit{iii}) log- and euclidean-based bias scores, where \textbf{lower} values indicate lower combined bias characteristics for both bias scores.}
    % % Model biases typically increase as you traverse \textit{down} the table.  All bias evaluations presented here are retrieved using the TBYB tool and code, with modified processes as described in Section 3.}
\end{table}

% We present a qualitative overview of how model bias characteristics manifest in the output space of different models in Fig. \ref{qual_fig}, using examples of models that exhibit significantly biased and unbiased behavior as reported in Table \ref{full_results_TABLE}. Through these results, we can see that the quantitative metrics are supported by qualitative findings. Models that report a higher average $M_G$ will generally demonstrate a greater semantic misalignment such that the output image is not relevant to the user input prompt as shown in the \textit{lambdalabs/sd-pokemon-diffusers} case. Models that report a low $B_D$ value will typically have constrained output representations with limited generative diversity, or a bias toward some concepts over others. Changes in $H_J$ are the easiest to interpret, showing input vs. generated object disparity.

Figure \ref{qual_fig} presents a qualitative overview of bias manifestations in model outputs, using examples from Table \ref{full_results_TABLE} to contrast biased and unbiased behaviors. These results align with quantitative metrics: a higher average $M_G$ will generally demonstrate a greater semantic misalignment (e.g. \textit{lambdalabs/sd-pokemon-diffusers}). Low $B_D$ models show constrained diversity or representational bias. Changes in $H_J$ are straightforward, reflecting disparities between input and generated objects.

The varying scales of the three metrics necessitate a logarithmic scale for comparing overall model bias. Each metric uniquely characterizes bias. Table \ref{full_results_TABLE} shows that low-bias models ($\downarrow\mathcal{B}{\log}$) typically report $B_D\geq14.0$, indicating a fairer distribution of generated objects. In contrast, highly biased models ($\mathcal{B}{\log}\geq-1.0$) report $B_D\leq7.0$, which suggests outliers or peaks in the output distribution.
For $H_J$, T2I models inherently hallucinate due to their semantically rich latent spaces. The average $H_J\approx0.55$ implies a 55\% IoU between prompted and generated objects. Foundation and photo-realism models cluster near the mean, whereas highly biased models exhibit extreme values, with a maximum $H_J=0.9721$ meaning just 2.79\% correlation between the input and output for the biased model.
$M_G$ remains low across most models, with a mean ($M_G=0.0333$) near the minimum ($M_G=0.0000$), indicating valid outputs $\approx$97\% of the time despite hallucinations. Models with high $M_G$ ($\geq0.60$) exhibit misaligned behavior, which based on model design may be intentional.

% These quantitative findings are further supported by the visualizations in Fig. \ref{qual_fig}.

% i.e., a higher disparity between input concepts vs. generated concepts will result in high observed $H_J$ values. 
% Combining these metrics into a single, log-based bias evaluation metric $\mathcal{B}_{\log}$ allows for intuitive analysis of model bias behavior.
% and a significant improvement over \cite{Vice2023B} which we hope can inspire continued research in this domain.  

\noindent\textbf{Evolution of Biases over Time}.
The release of the seminal latent diffusion work \citep{Rombach2022}, culminating in the public availability of the popular stable diffusion architecture on August 22, 2022, marked a pivotal moment in text-to-image generative models. Its launch on the HuggingFace Hub and subsequent community engagement spurred significant advancements in foundation models and task-specific variants. Accordingly, we use August 2022 as the starting point for our time-based analyses, with the latest evaluated model released in December 2024.

\begin{figure}
    \centering
    \includegraphics[width=0.95\linewidth]{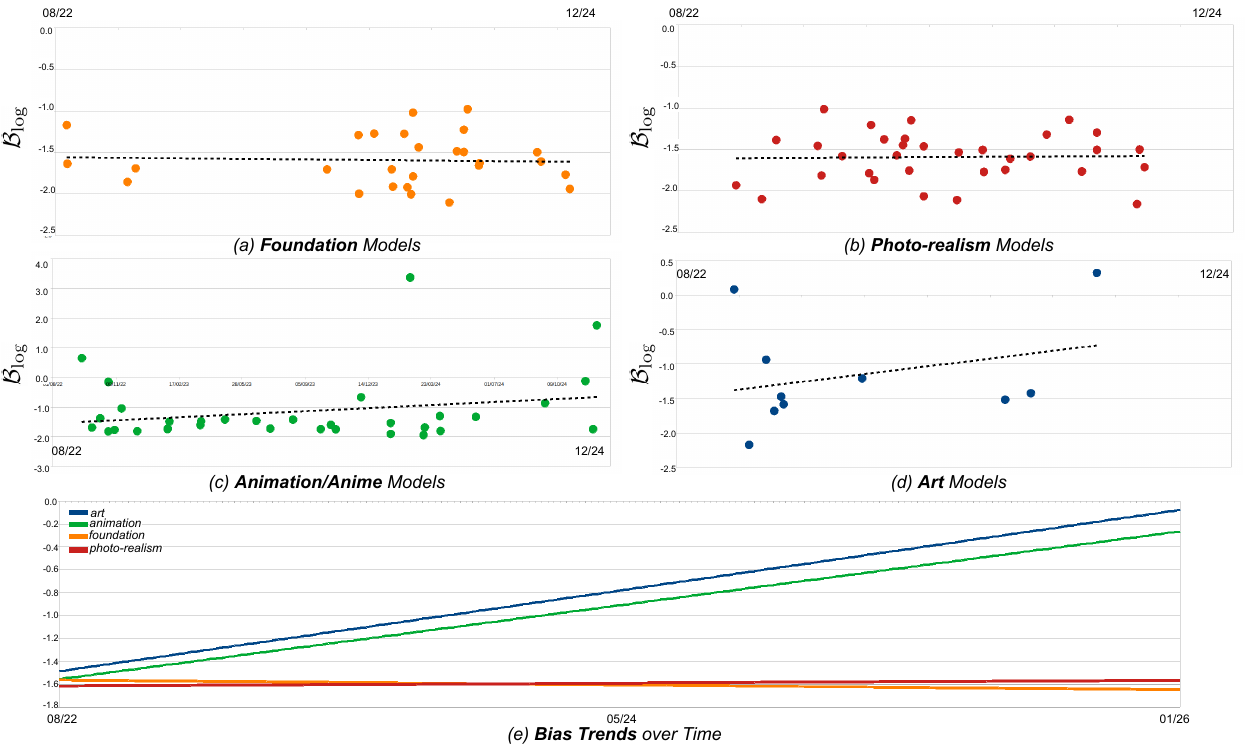}
    \vspace{-2mm}
    \caption{
    % Visualizing categorized $\mathcal{B}_{\log}$ model biases over time. \textit{(a)} Foundation models, \textit{(b)} Models trained for enhanced photo-realism, \textit{(c)} Models trained to copy anime/animation styles, \textit{(d)} \textit{non}-anime-inspired art generation models. All $x$ axes range between 08/2022 and 12/2024. Dotted lines in each sub-figure represent the linear trend, which we highlight (and extrapolate to 01/2026) in \textit{(e)}. }
    % Categorized $\mathcal{B}_{\log}$ model biases over time: \textit{(a)} Foundation models, \textit{(b)} Photo-realism models, \textit{(c)} Anime/animation-style models, \textit{(d)} Non-anime art models. X-axes span 08/2022$\rightarrow$12/2024. Dotted lines show linear trends, highlighted (and extrapolated to 01/2026) in \textit{(e)}.}
    Categorized temporal trends in $\mathcal{B}_{\log}$ model biases, spanning from 08/2022 $\rightarrow$ 12/2024. Dotted lines indicate linear trends, highlighted (and extrapolated to 01/2026) in \textit{(e)}.}
    \label{time_sorted_bias}
    \vspace{-4mm}
\end{figure}

Our evaluation spans 103 models over 28 months, presenting time-based bias analyses by individual metrics (Fig. \ref{metrics_fig}) and model categories (Fig. \ref{time_sorted_bias}). The timeline ($08/22 \rightarrow 12/24$) is consistent across sub-figures, with models grouped by task categories to examine trends. Bias trends, such as the steep increase in art and animation models’ bias over time (Fig. \ref{time_sorted_bias}\textit{(e)}), highlight the impact of hobbyists and practitioners embedding stylistic preferences or specific characters into these models. These intentional biases are reflected in their outputs, as supported by observations in Fig. \ref{metrics_fig}.
% resulting in high bias behavior.
% If these trends continue, we can expect the release of more significantly-biased models that cater to a variety of animation/art styles and fictional characters.

In comparison, we see that models associated with general tasks i.e., those that belong to the foundation and photo-realism model categories, have maintained consistent if not lower bias characteristics over time, as is the case with foundation models (See Fig. \ref{metrics_fig} and \ref{time_sorted_bias}\textit{(e)}). 
% Interestingly for foundation models, we can associate this with 
The increase in training data sizes and conscious improvements made to human-labeling and captioning in general have resulted in wider and denser manifolds with a greater diversity of concept representations. Significantly, looking at Table \ref{full_results_TABLE}, if we compare stable diffusion v1.4/2.1/3.5 rows, we can see that hallucination and distribution bias scores improve with each significant version upgrade through time.
% highlighting how the learned manifolds of these models have changed over time to mitigate the fundamental biases that were present in earlier versions.

\begin{table}[]

    \vspace{-4mm}
    \centering
    \caption{Observed mean and (standard deviation) across \textbf{model categories}. Column-wise bold values indicate the most biased behavior. We sort this table along the $\mathcal{S}_{pop.}$ column in descending order. Arrows in each column indicate the direction in which observed biases increases. }
    \label{cat_popularity_table}
    \vspace{2mm}
    \resizebox{0.85\linewidth}{!}{\begin{tabular}{l|ccccc}
         Model Type & $\mathcal{S}_{pop.}$ & $\downarrow$ $B_D$ & $\uparrow$ $H_J$ & $\uparrow$ $M_G$ & $\uparrow$ $\mathcal{B}_{\log}$\\
         \hline
         Foundation      & 7.2033 \tiny{($\pm$1.670)} & 10.7919 \tiny{($\pm$3.195)}          & 0.5175 \tiny{($\pm$0.038)}          & 0.0019 \tiny{($\pm$0.005)}          & -1.5960 \tiny{($\pm$0.308)} \\ 
         Photo Realism   & 6.4354 \tiny{($\pm$1.612)} & 11.1097 \tiny{($\pm$2.979)}          & 0.5392 \tiny{($\pm$0.030)}          & 0.0004 \tiny{($\pm$0.001)}          & -1.5949 \tiny{($\pm$0.292)} \\ 
         Art             & 6.2180 \tiny{($\pm$1.341)} & \textbf{10.0088} \tiny{($\pm$4.135)} & \textbf{0.6159} \tiny{($\pm$0.101)} & 0.0537 \tiny{($\pm$0.107)}          & -1.1581 \tiny{($\pm$0.786)} \\ 
         Animation/Anime & 6.0786 \tiny{($\pm$1.367)} & 10.4494 \tiny{($\pm$3.147)}          & 0.5969 \tiny{($\pm$0.120)}          & \textbf{0.0830} \tiny{($\pm$0.201)} & \textbf{-1.1503} \tiny{($\pm$1.134)} \\     
         \hline
    \end{tabular}}
    \vspace{-4mm}

\end{table}
\noindent\textbf{On the Influence of Model Type and Popularity}.
We conducted an evaluation of biases w.r.t. model categories and their popularity, exploiting Eq. (6) to quantify the latter. We report the results of these findings in Table \ref{cat_popularity_table}
and observe that foundation and photo-realism models are on-average, the most popular for users. Interestingly, these models tend to have more unbiased output representations when we consider the quantitative findings. Additionally, by analyzing the $\mathcal{B}_{\log}$ standard deviation results in Table \ref{cat_popularity_table},
% and Fig. \ref{statistics}\textit{(a)},
we see that foundation and photo-realism model performances are typically more consistent than art/animation counterparts. 

\noindent\textbf{De-noising Scheduler-Dependent Bias Evaluation}.
Much of the conditional latent diffusion process is predicated on the deployed de-noising scheduler. While similarities exist across different families of schedulers and the task remains the same i.e. use a conditional vector to guide latent de-noising steps to generate an aligned image representation of the input prompt, the mathematical foundations of each scheduler is unique. We report the descriptive statistics of different schedulers in Table \ref{scheduler_table}, highlighting eight scheduler categories. The \textit{FlowMatchEulerDiscrete} scheduler is deployed in Stable diffusion 3 variants which points to its high popularity and low $\mathcal{B}_{\log}$ score. Recently, flow-based de-noising schedulers have gained increased attention in state-of-the-art T2I models like Stable Diffusion 3 and FLUX \citep{Esser2024, FLUXAI2024}.

In comparison, the EulerDiscrete scheduler \citep{Karras2022} reports the largest bias and the highest average miss-rate. Incremental improvements in scheduler architectures and technology since the release of EulerDiscrete, along with the modern T2I models opting for modern schedulers are logical reasons as to why this scheduler reports significantly high bias scores. Similarly, the \textit{EulerAncestralDiscrete} scheduler, which contributes ``ancestral sampling'' also boasts consistent performance with its predecessor. These seminal works have inspired improvements which as shown through the \textit{FlowMatchEulerDiscrete} scheduler, have resulted in significant performance gains.
% over time.

% Ultimately, the scheduler is a critical component in the T2I generative pipeline and if we can understand any potential underlying bias characteristics of different schedulers, this may provide users and AI practitioners with additional insights into the construction of these complex, multimodal generative pipelines. 
We note that while using quantifiable metrics like those reported here present a step in the right direction, any definitive correlations will require a deeper analysis into the schedulers themselves.

\begin{table}[]
    \centering
    \caption{Observed mean and (standard deviation) across deployed \textbf{schedulers}. Column-wise bold values indicate the most biased behavior. We sort the schedulers along the $\mathcal{S}_{pop.}$ in descending order. Arrows in each column indicate the direction in which observed bias increases. }
    \label{scheduler_table}
    \vspace{2mm}
    \resizebox{0.85\linewidth}{!}{\begin{tabular}{l|ccccc}
         Scheduler & $\mathcal{S}_{pop.}$ & $\downarrow$ $B_D$ & $\uparrow$ $H_J$ & $\uparrow$ $M_G$ & $\uparrow$ $\mathcal{B}_{\log}$\\
         \hline
         FlowMatchEulerDiscrete  & 7.5101 \tiny{($\pm$2.181)} & 12.2788 \tiny{($\pm$1.541)}         & 0.4959 \tiny{($\pm$0.008)}          & 0.0000 \tiny{($\pm$0.000)}          & -1.8177 \tiny{($\pm$0.106)} \\ 
         KDPM2AncestralDiscrete  & 7.4314 \tiny{($\pm$0.424)} & 9.5803 \tiny{($\pm$2.667)}          & 0.5279 \tiny{($\pm$0.010)}          & 0.0000 \tiny{($\pm$0.000)}          & -1.4894 \tiny{($\pm$0.304)} \\ 
         DDIM/DDPM               & 7.3371 \tiny{($\pm$1.973)} & 10.1441 \tiny{($\pm$2.548)}         & 0.5321 \tiny{($\pm$0.039)}          & 0.0067 \tiny{($\pm$0.014)}           & -1.5185 \tiny{($\pm$0.261)} \\ 
         EulerAncestralDiscrete  & 7.1003 \tiny{($\pm$1.667)} & \textbf{9.1328} \tiny{($\pm$3.083)} & 0.5447 \tiny{($\pm$0.084)}          & 0.0213 \tiny{($\pm$0.060)}          & -1.3332 \tiny{($\pm$0.552)}\\ 
         DEISMultistep           & 6.6288 \tiny{($\pm$0.357)} & 12.8746 \tiny{($\pm$3.953)}         & \textbf{0.5713} \tiny{($\pm$0.025)} & 0.0067 \tiny{($\pm$0.012)}          & -1.6710 \tiny{($\pm$0.341)} \\ 
         PNDM                    & 6.3172 \tiny{($\pm$1.584)} & 11.1714 \tiny{($\pm$3.002)}         & 0.5694 \tiny{($\pm$0.072)}          & 0.0302 \tiny{($\pm$0.106)}          & -1.4658 \tiny{($\pm$0.547)} \\ 
         EulerDiscrete           & 6.2381 \tiny{($\pm$1.392)} & 10.3751 \tiny{($\pm$3.573)}         & 0.5702 \tiny{($\pm$0.124)}          & \textbf{0.0621} \tiny{($\pm$0.190)} & \textbf{-1.2184} \tiny{($\pm$1.175)} \\ 
         DPMSolverMultistep      & 4.9966 \tiny{($\pm$0.603)} & 12.1618 \tiny{($\pm$3.654)}         & 0.5662 \tiny{($\pm$0.046)}          & 0.0042 \tiny{($\pm$0.008)}          & -1.6255 \tiny{($\pm$0.360)} \\          
         % misc                             & \textbf{7.7674} & 8.4057          & 0.5026 & 0.0022 & -1.3638 \\ 
         \hline
    \end{tabular}}
    \vspace{-4mm}
    % \vspace{-2mm}

\end{table}

% \subsection{Geopolitical Task-oriented evaluation}

%% file: sections/5_conclusion.tex
\vspace{-2mm}
\section{Conclusion}
\vspace{-2mm}
We have conducted an extensive evaluation of text-to-image models, utilizing the open HuggingFace Hub to facilitate our analyses of the bias characteristics of 103 unique models. To improve on existing evaluation methodologies, we combine three independent metrics i.e., (i) distribution bias, (ii) Jaccard hallucination and, (iii) generative miss-rate into a single log-scaled metric. By accounting for various generative model categories and quantifying public engagement, we have presented a comprehensive set of model evaluations. Identifying the fundamental bias characteristics of large, publicly-available text-to-image models is a critical task that must be considered in a democratized AI environment, considering the exposure of these models to wider audiences that continues to grow over time. 
So, to answer the question of ``\textit{are models more biased now than they were 3 years ago?}'' really depends on the task. We see that iterative releases of Stable diffusion models for example, have resulted in marginal improvements in bias characteristics over time (from SD 1.1 to 3.5).  Foundation and photo-realism models have demonstrated significant reductions in hallucination and increases in alignment, which is beneficial for improving the reliability for a wider range of audiences. As it pertains to style-transferred, art and animation models, these have demonstrated increased bias characteristics - a byproduct of intentionally-designing models to achieve specific tasks. We hope this work inspires further research in the field and a greater exposure to bias evaluation efforts.
% Through general and task-oriented evaluations, we identify fundamental bias characteristics of evaluated models provide a geopolitical task to demonstrate different bias directions. 
% ke TBYB application.
\vspace{-2mm}
\section*{Acknowledgments}
\vspace{-2mm}
This research and Dr. Jordan Vice are supported by the NISDRG project \#20100007, funded by the Australian Government. Dr. Naveed Akhtar is a recipient of the ARC Discovery Early Career Researcher Award (project \#DE230101058), funded by the Australian Government. Professor Ajmal Mian is the recipient of an ARC Future Fellowship Award (project \#FT210100268) funded by the Australian Government.

%% file: sections/X_supplementary.tex
\clearpage
\setcounter{table}{0}
\renewcommand{\thetable}{A\arabic{table}}
\section*{Appendices}
\textbf{Appendix A:} Full bias evaluations results of 103 text-to-image generative models. Evaluations are reported in $B_{\log}$ ascending order. Truncated results in Table \ref{full_results_TABLE} of the main manuscript are a sub-set of the full results presented here.
\begin{table}[h]
% \vspace{-6mm}
    \centering
    % \caption{Full bias evaluations results, sorted in $B_{\log}$ ascending order. Truncated evaluation results are presented in Table \ref{full_results_TABLE}.}
    \label{appendix_A}
    \vspace{2mm}
    \resizebox{\linewidth}{!}{\begin{tabular}{ll|lcc|ccccc}
         Model & Task Category & Denoiser & Resolution & Release (dd/mm/yy) & $\mathcal{S}_{pop.}$&$B_D$ & $H_J$ & $M_G$ & $B_{\log}$\\
         \hline
            Envvi/Inkpunk-Diffusion & art & PNDMScheduler & 512 x 512 & 25/11/22 & 7.2323 & 18.9000 & 0.5346 & 0.0033 & -2.1711 \\ 
            Yntec/beLIEve & photo realism & DPMSolverMultistepScheduler & 768 x 768 & 01/08/24 & 5.2547 & 17.6176 & 0.5083 & 0.0000 & -2.1589 \\ 
            segmind/SSD-1B & photo realism & EulerDiscreteScheduler & 1024 x 1024 & 19/10/23 & 6.7116 & 15.7000 & 0.4747 & 0.0000 & -2.1098 \\ 
            RunDiffusion/Juggernaut-X-v10 & foundation & EulerDiscreteScheduler & 1024 x 1024 & 20/04/24 & 6.8125 & 16.3571 & 0.4992 & 0.0000 & -2.1031 \\ 
            prompthero/openjourney-v4 & photo realism & PNDMScheduler & 512 x 512 & 12/12/22 & 8.4414 & 15.9211 & 0.4881 & 0.0000 & -2.0981 \\ 
            Lykon/dreamshaper-8 & photo realism & DEISMultistepScheduler & 512 x 512 & 27/08/23 & 6.8769 & 17.3947 & 0.5467 & 0.0000 & -2.0649 \\ 
            RunDiffusion/Juggernaut-XL-v9 & foundation & DDPMScheduler & 1024 x 1024 & 19/02/24 & 8.6025 & 14.4048 & 0.4847 & 0.0000 & -2.0046 \\ 
            stabilityai/sd-turbo & foundation & EulerDiscreteScheduler & 512 x 512 & 28/11/23 & 7.9498 & 14.5476 & 0.4930 & 0.0000 & -1.9982 \\ 
            eienmojiki/Anything-XL & animation / anime & EulerAncestralDiscreteScheduler & 1024 x 1024 & 11/03/24 & 5.6000 & 14.8333 & 0.5287 & 0.0000 & -1.9446 \\ 
            stabilityai/stable-diffusion-3.5-medium & foundation & FlowMatchEulerDiscreteScheduler & 1024 x 1024 & 29/10/24 & 8.2481 & 14.0455 & 0.5049 & 0.0000 & -1.9393 \\ 
            MirageML/dreambooth-nike & photo realism & PNDMScheduler & 512 x 512 & 01/11/22 & 3.3402 & 14.4048 & 0.5206 & 0.0000 & -1.9323 \\ 
            dataautogpt3/ProteusV0.3 & foundation & EulerDiscreteScheduler & 1024 x 1024 & 13/02/24 & 7.3949 & 14.5833 & 0.5324 & 0.0000 & -1.9196 \\ 
            stablediffusionapi/juggernaut-reborn & foundation & PNDMScheduler & 512 x 512 & 21/01/24 & 5.0168 & 13.9783 & 0.5073 & 0.0167 & -1.9129 \\ 
            hongdthaui/ManmaruMix\_v30 & animation / anime & PNDMScheduler & 512 x 512 & 19/01/24 & 4.4554 & 16.2391 & 0.5857 & 0.0033 & -1.9029 \\ 
            digiplay/Photon\_v1 & photo realism & EulerDiscreteScheduler & 768 x 768 & 09/06/23 & 6.6038 & 13.4583 & 0.5195 & 0.0000 & -1.8667 \\ 
            lambdalabs/miniSD-diffusers & foundation & PNDMScheduler & 512 x 512 & 24/11/22 & 5.5457 & 14.6600 & 0.5640 & 0.0000 & -1.8550 \\ 
            nitrosocke/mo-di-diffusion & animation / anime & PNDMScheduler & 512 x 512 & 28/10/22 & 7.4843 & 14.3500 & 0.5696 & 0.0033 & -1.8175 \\ 
            Lykon/DreamShaper & photo realism & PNDMScheduler & 512 x 512 & 17/03/23 & 9.2373 & 13.7000 & 0.5521 & 0.0000 & -1.8142 \\ 
            ItsJayQz/GTA5\_Artwork\_Diffusion & animation / anime & PNDMScheduler & 512 x 512 & 13/12/22 & 6.9009 & 12.9348 & 0.5257 & 0.0033 & -1.8106 \\ 
            digiplay/PerfectDeliberate-Anime\_v2  & animation / anime & EulerDiscreteScheduler & 768 x 768 & 07/04/24 & 5.9243 & 14.6923 & 0.5863 & 0.0000 & -1.8048 \\ 
            RunDiffusion/Juggernaut-XL-v6 & foundation & EulerDiscreteScheduler & 1024 x 1024 & 22/02/24 & 5.9995 & 14.1296 & 0.5766 & 0.0000 & -1.7889 \\ 
            Lykon/AbsoluteReality & photo realism & PNDMScheduler & 512 x 512 & 01/06/23 & 5.3458 & 12.6923 & 0.5297 & 0.0000 & -1.7866 \\ 
            segmind/Segmind-Vega & photo realism & EulerDiscreteScheduler & 1024 x 1024 & 01/12/23 & 6.1051 & 12.8462 & 0.5427 & 0.0000 & -1.7706 \\ 
            nitrosocke/redshift-diffusion & animation / anime & PNDMScheduler & 512 x 512 & 07/11/22 & 6.5239 & 12.0769 & 0.5139 & 0.0000 & -1.7699 \\ 
            stabilityai/stable-diffusion-3.5-large & foundation & FlowMatchEulerDiscreteScheduler & 1024 x 1024 & 22/10/24 & 9.2260 & 11.5769 & 0.4939 & 0.0000 & -1.7680 \\ 
            circulus/canvers-real-v3.9.1 & photo realism & PNDMScheduler & 512 x 512 & 05/05/24 & 3.8623 & 12.4615 & 0.5308 & 0.0000 & -1.7658 \\ 
            digiplay/AbsoluteReality\_v1.8.1 & photo realism & DDIMScheduler & 768 x 768 & 04/08/23 & 6.5521 & 11.6481 & 0.5034 & 0.0000 & -1.7552 \\ 
            Yntec/YiffyMix & animation / anime & PNDMScheduler & 512 x 512 & 24/10/23 & 5.9138 & 12.1296 & 0.5259 & 0.0000 & -1.7492 \\ 
            Yntec/RealLife & photo realism & EulerDiscreteScheduler & 768 x 768 & 04/01/24 & 5.5369 & 11.9828 & 0.5216 & 0.0000 & -1.7461 \\ 
            digiplay/MilkyWonderland\_v1 & animation / anime & EulerDiscreteScheduler & 768 x 768 & 30/09/23 & 4.8211 & 12.6538 & 0.5394 & 0.0167 & -1.7458 \\ 
            aipicasso/emi-3 & animation / anime & FlowMatchEulerDiscreteScheduler & 1024 x 1024 & 05/12/24 & 5.0561 & 11.2140 & 0.4891 & 0.0000 & -1.7457 \\ 
            WarriorMama777/AbyssOrangeMix2 & animation / anime & PNDMScheduler & 512 x 512 & 30/01/23 & 7.8778 & 12.3400 & 0.5353 & 0.0067 & -1.7397 \\ 
            WarriorMama777/AbyssOrangeMix & animation / anime & PNDMScheduler & 512 x 512 & 30/01/23 & 7.8316 & 11.5000 & 0.5096 & 0.0000 & -1.7299 \\ 
            liamhvn/disney-pixar-cartoon-b & animation / anime & PNDMScheduler & 512 x 512 & 12/07/23 & 4.9147 & 12.8548 & 0.5567 & 0.0167 & -1.7235 \\ 
            openart-custom/CrystalClearXL & photo realism & EulerDiscreteScheduler & 1024 x 1024 & 13/08/24 & 5.5520 & 12.2308 & 0.5464 & 0.0000 & -1.7134 \\ 
            stablediffusionapi/sdxl-unstable-diffusers-y & foundation & EulerDiscreteScheduler & 1024 x 1024 & 08/10/23 & 5.1028 & 11.3462 & 0.5151 & 0.0000 & -1.7052 \\ 
            dataautogpt3/ProteusV0.2 & foundation & KDPM2AncestralDiscreteScheduler & 1024 x 1024 & 19/01/24 & 7.1317 & 11.4655 & 0.5207 & 0.0000 & -1.7040 \\ 
            stabilityai/stable-diffusion-2-1 & foundation & DDIMScheduler & 512 x 512 & 07/12/22 & 10.5860 & 11.7963 & 0.5349 & 0.0100 & -1.6921 \\ 
            nitrosocke/Arcane-Diffusion & animation / anime & LMSDiscreteScheduler & 512 x 512 & 02/10/22 & 7.1673 & 10.9074 & 0.5004 & 0.0067 & -1.6887 \\ 
            cagliostrolab/animagine-xl-3.1 & animation / anime & EulerAncestralDiscreteScheduler & 1024 x 1024 & 13/03/24 & 8.9596 & 11.2143 & 0.5179 & 0.0000 & -1.6876 \\ 
            nuigurumi/basil\_mix & art & PNDMScheduler & 512 x 512 & 04/01/23 & 7.0493 & 12.1800 & 0.5598 & 0.0000 & -1.6792 \\ 
            fluently/Fluently-XL-Final & foundation & EulerAncestralDiscreteScheduler & 1024 x 1024 & 06/06/24 & 6.6590 & 10.8226 & 0.5147 & 0.0000 & -1.6586 \\ 
            CompVis/stable-diffusion-v1-4 & foundation & PNDMScheduler & 512 x 512 & 20/08/22 & 10.7885 & 11.7258 & 0.5621 & 0.0000 & -1.6360 \\ 
            SPO-Diffusion-Models/SPO-SDXL\_4k-p\_10ep & foundation & EulerDiscreteScheduler & 1024 x 1024 & 07/06/24 & 6.2646 & 11.1667 & 0.5426 & 0.0000 & -1.6307 \\ 
            krnl/realisticVisionV51\_v51VAE & photo realism & PNDMScheduler & 512 x 512 & 12/01/24 & 5.3375 & 11.1667 & 0.5510 & 0.0000 & -1.6122 \\ 
            openart-custom/AlbedoBase & foundation & EulerDiscreteScheduler & 1024 x 1024 & 13/09/24 & 5.5861 & 11.0161 & 0.5462 & 0.0000 & -1.6093 \\ 
            xyn-ai/anything-v4.0 & animation / anime & PNDMScheduler & 512 x 512 & 23/03/23 & 6.4597 & 11.6250 & 0.5692 & 0.0067 & -1.6044 \\ 
            lemon2431/toonify\_v20 & animation / anime & PNDMScheduler & 512 x 512 & 16/10/23 & 4.1148 & 10.9412 & 0.5469 & 0.0033 & -1.5976 \\ 
            SG161222/RealVisXL\_V4.0 & photo realism & EulerDiscreteScheduler & 1024 x 1024 & 13/02/24 & 8.6080 & 10.6250 & 0.5405 & 0.0000 & -1.5856 \\ 
            ckpt/anything-v4.5 & art & PNDMScheduler & 512 x 512 & 19/01/23 & 5.8262 & 11.5968 & 0.5784 & 0.0033 & -1.5837 \\ 
            emilianJR/chilloutmix\_NiPrunedFp32Fix & photo realism & PNDMScheduler & 512 x 512 & 19/04/23 & 6.0849 & 10.7941 & 0.5498 & 0.0000 & -1.5810 \\ 
            Kernel/sd-nsfw & photo realism & PNDMScheduler & 512 x 512 & 15/07/23 & 5.5154 & 11.5333 & 0.5828 & 0.0000 & -1.5711 \\ 
            Lykon/AAM\_XL\_AnimeMix & animation / anime & EulerDiscreteScheduler & 1024 x 1024 & 19/01/24 & 6.8090 & 9.0152 & 0.4843 & 0.0000 & -1.5367 \\ 
            stablediffusionapi/realistic-stock-photo & photo realism & EulerDiscreteScheduler & 1024 x 1024 & 22/10/23 & 5.1367 & 10.2188 & 0.5451 & 0.0000 & -1.5366 \\ 
            GraydientPlatformAPI/comicbabes2 & art & PNDMScheduler & 512 x 512 & 07/01/24 & 4.0530 & 10.0313 & 0.5447 & 0.0033 & -1.5156 \\ 
            SG161222/Realistic\_Vision\_V6.0\_B1\_noVAE & photo realism & PNDMScheduler & 896 x 896 & 29/11/23 & 7.7733 & 11.1571 & 0.5957 & 0.0000 & -1.5066 \\ 
            scenario-labs/juggernaut\_reborn & photo realism & DPMSolverMultistepScheduler & 512 x 512 & 29/05/24 & 4.5414 & 10.0294 & 0.5504 & 0.0000 & -1.5061 \\ 
            SG161222/RealVisXL\_V5.0 & photo realism & DDIMScheduler & 1024 x 1024 & 05/08/24 & 6.6694 & 9.9412 & 0.5482 & 0.0000 & -1.5022 \\ 
            stable-diffusion-v1-5/stable-diffusion-v1-5 & foundation & PNDMScheduler & 512 x 512 & 07/09/24 & 9.2497 & 9.6429 & 0.5361 & 0.0000 & -1.4981 \\ 
            Corcelio/openvision & foundation & EulerDiscreteScheduler & 1024 x 1024 & 13/05/24 & 5.7229 & 9.2576 & 0.5161 & 0.0033 & -1.4963 \\ 
            gsdf/Counterfeit-V2.5 & animation / anime & DDIMScheduler & 512 x 512 & 02/02/23 & 8.8241 & 12.2059 & 0.6204 & 0.0433 & -1.4890 \\ 
            fluently/Fluently-XL-v4 & foundation & EulerAncestralDiscreteScheduler & 1024 x 1024 & 02/05/24 & 6.8337 & 9.2222 & 0.5205 & 0.0000 & -1.4866 \\ 
            Ojimi/anime-kawai-diffusion & animation / anime & DEISMultistepScheduler & 512 x 512 & 24/03/23 & 6.2192 & 11.1667 & 0.5968 & 0.0200 & -1.4843 \\ 
            tilake/China-Chic-illustration & art & PNDMScheduler & 512 x 512 & 15/01/23 & 5.7534 & 10.0294 & 0.5655 & 0.0000 & -1.4720 \\ 
            digiplay/FormCleansingMix\_v1 & animation / anime & DPMSolverMultistepScheduler & 768 x 768 & 20/06/23 & 4.4630 & 10.8243 & 0.5936 & 0.0167 & -1.4647 \\ 
            Lykon/dreamshaper-7 & photo realism & DEISMultistepScheduler & 512 x 512 & 27/08/23 & 6.7903 & 10.0625 & 0.5704 & 0.0000 & -1.4638 \\ 
            gligen/diffusers-generation-text-box & photo realism & PNDMScheduler & 512 x 512 & 11/03/23 & 5.5617 & 10.0294 & 0.5719 & 0.0000 & -1.4570 \\ 
            stabilityai/stable-diffusion-xl-base-1.0 & photo realism & EulerDiscreteScheduler & 1024 x 1024 & 25/07/23 & 11.1360 & 8.6515 & 0.5076 & 0.0000 & -1.4492 \\ 
            pt-sk/stable-diffusion-1.5 & foundation & PNDMScheduler & 512 x 512 & 02/03/24 & 5.0628 & 9.5556 & 0.5584 & 0.0000 & -1.4397 \\ 
            playgroundai/playground-v2.5-1024px-aesthetic & art & EDMDPMSolverMultistepScheduler & 1024 x 1024 & 17/02/24 & 8.8552 & 9.5270 & 0.5649 & 0.0000 & -1.4219 \\ 
            danbrown/RevAnimated-v1-2-2 & animation / anime & DDIMScheduler & 256 x 256 & 01/05/23 & 5.0277 & 8.5000 & 0.5134 & 0.0000 & -1.4198 \\ 
            redstonehero/animesh\_prunedv21 & animation / anime & PNDMScheduler & 512 x 512 & 17/08/23 & 4.0369 & 9.6622 & 0.5705 & 0.0033 & -1.4197 \\ 
            dreamlike-art/dreamlike-photoreal-2.0 & photo realism & DDIMScheduler & 768 x 768 & 04/01/23 & 8.6776 & 8.8415 & 0.5451 & 0.0033 & -1.3885 \\ 
            emilianJR/epiCRealism & photo realism & PNDMScheduler & 512 x 512 & 25/06/23 & 6.9505 & 8.8846 & 0.5499 & 0.0067 & -1.3793 \\ 
            naclbit/trinart\_characters\_19.2m\_stable\_diffusion\_v1 & animation / anime & PNDMScheduler & 512 x 512 & 15/10/22 & 6.2754 & 12.3750 & 0.6536 & 0.0767 & -1.3757 \\ 
            segmind/tiny-sd & photo realism & DPMSolverMultistepScheduler & 512 x 512 & 28/07/23 & 5.7272 & 10.1757 & 0.6124 & 0.0000 & -1.3722 \\ 
            yodayo-ai/kivotos-xl-2.0 & animation / anime & EulerAncestralDiscreteScheduler & 1024 x 1024 & 02/06/24 & 6.5836 & 7.1410 & 0.4717 & 0.0000 & -1.3277 \\ 
            ZB-Tech/Text-to-Image & photo realism & PNDMScheduler & 512 x 512 & 10/03/24 & 6.4625 & 8.7162 & 0.5697 & 0.0000 & -1.3220 \\ 
            digiplay/aurorafantasy\_v1 & animation / anime & EulerDiscreteScheduler & 768 x 768 & 06/04/24 & 5.2524 & 8.8864 & 0.5813 & 0.0133 & -1.3005 \\ 
            stablediffusionapi/mklan-xxx-nsfw-pony & photo realism & EulerDiscreteScheduler & 1024 x 1024 & 29/05/24 & 6.5259 & 7.4744 & 0.5100 & 0.0000 & -1.2981 \\ 
            stabilityai/sdxl-turbo & foundation & EulerAncestralDiscreteScheduler & 512 x 512 & 27/11/23 & 10.1726 & 8.0778 & 0.5493 & 0.0000 & -1.2922 \\ 
            Lykon/dreamshaper-xl-v2-turbo & foundation & EulerDiscreteScheduler & 1024 x 1024 & 08/02/24 & 6.6775 & 6.7381 & 0.4679 & 0.0000 & -1.2769 \\ 
            dataautogpt3/OpenDalleV1.1 & foundation & KDPM2AncestralDiscreteScheduler & 1024 x 1024 & 22/12/23 & 7.7311 & 7.6951 & 0.5351 & 0.0000 & -1.2747 \\ 
            Corcelio/mobius & foundation & EulerDiscreteScheduler & 1024 x 1024 & 13/05/24 & 6.0584 & 7.2273 & 0.5280 & 0.0000 & -1.2271 \\ 
            kandinsky-community/kandinsky-2-1 & art & DDIMScheduler & 512 x512 & 24/05/23 & 6.5896 & 7.1735 & 0.5332 & 0.0000 & -1.2086 \\ 
            friedrichor/stable-diffusion-2-1-realistic & photo realism & DDIMScheduler & 512 x 512 & 04/06/23 & 4.5053 & 6.7857 & 0.5061 & 0.0033 & -1.2060 \\ 
            CompVis/stable-diffusion-v1-1 & foundation & PNDMScheduler & 512 x 512 & 19/08/22 & 6.5538 & 6.8864 & 0.5218 & 0.0200 & -1.1716 \\ 
            stablediffusionapi/realistic-vision-51 & photo realism & PNDMScheduler & 512 x 512 & 07/08/23 & 5.8350 & 6.9490 & 0.5456 & 0.0000 & -1.1498 \\ 
            UnfilteredAI/NSFW-gen-v2 & photo realism & EulerAncestralDiscreteScheduler & 1024 x 1024 & 15/04/24 & 6.8121 & 6.4111 & 0.5102 & 0.0000 & -1.1442 \\ 
            nitrosocke/Ghibli-Diffusion & animation / anime & PNDMScheduler & 704 x 512 & 18/11/22 & 7.6340 & 6.3478 & 0.5523 & 0.0000 & -1.0444 \\ 
            dataautogpt3/ProteusV0.4-Lightning & foundation & EulerDiscreteScheduler & 1024 x 1024 & 22/02/24 & 6.2324 & 4.6739 & 0.4055 & 0.0000 & -1.0220 \\ 
            SG161222/Realistic\_Vision\_V2.0 & photo realism & PNDMScheduler & 512 x 512 & 21/03/23 & 8.3000 & 6.1154 & 0.5480 & 0.0000 & -1.0167 \\ 
            sd-community/sdxl-flash & foundation & DPMSolverSinglestepScheduler & 1024 x 1024 & 19/05/24 & 7.2798 & 4.7826 & 0.4424 & 0.0000 & -0.9809 \\ 
            Mitsua/mitsua-diffusion-cc0 & art & PNDMScheduler & 512 x 512 & 22/12/22 & 5.1305 & 10.8947 & 0.7426 & 0.0900 & -0.9368 \\ 
            OnomaAIResearch/Illustrious-xl-early-release-v0 & animation / anime & EulerDiscreteScheduler & 1024 x 1024 & 20/09/24 & 7.3851 & 11.6000 & 0.7524 & 0.1700 & -0.8688 \\ 
            digiplay/ZHMix-Dramatic-v2.0 & animation / anime & EulerDiscreteScheduler & 768 x 768 & 03/12/23 & 4.7306 & 5.7800 & 0.6506 & 0.0333 & -0.6689 \\ 
            DGSpitzer/Cyberpunk-Anime-Diffusion & animation / anime & PNDMScheduler & 704 x 704 & 28/10/22 & 6.7796 & 2.9722 & 0.5845 & 0.0600 & -0.1492 \\ 
            Emanon14/NONAMEmix\_v1 & animation / anime & EulerAncestralDiscreteScheduler & 1024 x 1024 & 23/11/24 & 5.1821 & 5.3400 & 0.7447 & 0.1700 & -0.1237 \\ 
            Onodofthenorth/SD\_PixelArt\_SpriteSheet\_Generator & art & PNDMScheduler & 512 x 512 & 01/11/22 & 6.4996 & 6.5930 & 0.7898 & 0.3367 & 0.0841 \\ 
            Niggendar/duchaitenPonyXLNo\_ponyNoScoreV40 & art & EulerDiscreteScheduler & 1024 x 1024 & 01/06/24 & 5.1912 & 3.1620 & 0.7458 & 0.1000 & 0.3237 \\ 
            lambdalabs/sd-pokemon-diffusers & animation / anime & PNDMScheduler & 512 x 512 & 16/09/22 & 6.3118 & 9.6509 & 0.8639 & 0.6033 & 0.6519 \\ 
            Raelina/Raehoshi-illust-XL-3 & animation / anime & EulerDiscreteScheduler & 1024 x 1024 & 11/12/24 & 3.7427 & 4.8772 & 0.8926 & 0.6700 & 1.7549 \\ 
            monadical-labs/minecraft-skin-generator-sdxl & animation / anime & EulerDiscreteScheduler & 768 x 768 & 19/02/24 & 5.3317 & 5.9786 & 0.9721 & 0.7933 & 3.3680 \\ 
         \hline
    \end{tabular}}
    \vspace{-2mm}
    % \vspace{-2mm}

    % A full list of tabulated results is reported in the supplementary material.}
    % Our evaluation results include: (\textit{i}) popularity score `$\mathcal{S}_{pop.}$' which is derived from the number of likes and downloads of the model, (\textit{ii}) the bias evaluation metrics introduced in \cite{Vice2023B} and, (\textit{iii}) log- and euclidean-based bias scores, where \textbf{lower} values indicate lower combined bias characteristics for both bias scores.}
    % % Model biases typically increase as you traverse \textit{down} the table.  All bias evaluations presented here are retrieved using the TBYB tool and code, with modified processes as described in Section 3.}
\end{table}